# Biomimicry in Radiation Therapy: Optimizing Patient Scheduling for Improved Treatment Outcomes

Keshav Kumar K.[1*], Dr. NVSL Narasimham[2]

[1]Department of Mathematics, G. Narayanamma Institute of Technology and Science (for Women), Hyderabad-500 104, Telangana State, India.
*Orcid ID:* https://orcid.org/0000-0002-9211-2960

[2]Department of Mathematics, G. Narayanamma Institute of Technology and Science (for Women), Hyderabad-500 104, Telangana State, India.

*Corresponding Author: keshav.gnits@gmail.com

*Abstract—* In the realm of medical science, the pursuit of enhancing treatment efficacy and patient outcomes continues to drive innovation. This study delves into the integration of biomimicry principles within the domain of Radiation Therapy (RT) to optimize patient scheduling, ultimately aiming to augment treatment results. RT stands as a vital medical technique for eradicating cancer cells and diminishing tumor sizes. Yet, the manual scheduling of patients for RT proves both laborious and intricate. In this research, the focus is on automating patient scheduling for RT through the application of optimization methodologies. Three bio-inspired algorithms are employed for optimization to tackle the complex online stochastic scheduling problem. These algorithms include the Genetic Algorithm (GA), Firefly Optimization (FFO), and Wolf Optimization (WO). These algorithms are harnessed to address the intricate challenges of online stochastic scheduling. Through rigorous evaluation, involving the scrutiny of convergence time, runtime, and objective values, the comparative performance of these algorithms is determined. The results of this study unveil the effectiveness of the applied bio-inspired algorithms in optimizing patient scheduling for RT. Among the algorithms examined, WO emerges as the frontrunner, consistently delivering superior outcomes across various evaluation criteria. The optimization approach showcased in this study holds the potential to streamline processes, reduce manual intervention, and ultimately improve treatment outcomes for patients undergoing RT.

*Keywords— Radiotherapy, Optimization, Runtime, Convergence, Schedule, Stochastic Function.*

## I. Introduction

The predicted global incidence of new cancer cases per year is set to surpass 27 million by 2040, which is 1.5 times higher than the estimated count of 18.1 million cases in 2018, as outlined in the 2020 world cancer report [1]. This significant increase, coupled with the growing need to optimize healthcare expenditure, places pressure on healthcare facilities like radiotherapy centers to enhance their operational efficiency [2].

RT, commonly known as radiotherapy, is a frequently employed approach for treating cancer patients, aiming to eliminate cancerous cells while preserving nearby healthy tissue. It is often used in conjunction with or as an alternative to surgery and chemotherapy. Typically, radiotherapy appointments are scheduled several days or weeks ahead, with emergencies necessitating immediate treatment being rare. However, real-world data shows substantial variability in the duration of radiotherapy sessions, despite accurate predictions being feasible during the meticulous treatment planning phase by medical physicists. The presence of uncertainty poses a significant challenge in the scheduling process [3].

Although advancements in technology have led to improved precision in radiotherapy equipment, unfortunately, the scheduling procedures have not advanced at the same rate. Many medical institutions focused on cancer treatment still rely on manual and fragmented scheduling approaches. This results in extended wait times for patients, specifically the delay between requesting an appointment and undergoing treatment. This delay, known as access delay time, negatively impacts patient outcomes and recovery. As a response, several authors [4-6] emphasize the critical nature of this issue and suggest different scheduling protocols. Research has affirmed that delays in RT amplify the risk of patient mortality [7]. As a result, certain medical establishments have chosen to enhance the frequency of shifts or procure additional equipment in order to provide more effective medical attention to patients. Nevertheless, these approaches come with significant financial expenses.

This study was conducted to enhance scheduling efficiency, enabling medical institutions to provide more streamlined patient care without substantial investments. This paper's main objective is to address the issue of patient scheduling in RT through the introduction of a mathematical model. The goal is to reduce the amount of time patients spend waiting, which includes the duration between the decision to arrange a treatment session and the actual administration of the treatment, often known as the access time.

## II. LITERATURE SURVEY

In the study presented in [8], a novel approach employing Column Generation (CG) is introduced to address the RT patient scheduling challenge. The suggested model incorporates real-world restrictions such as patient-specific protocols, machine compatibilities, and numerous hospital sites, all of which are essential for efficient schedule development. This model notably incorporates planned treatment interruptions due to equipment maintenance, a crucial consideration in patient scheduling due to its potential impact on workflow bottlenecks. Various techniques, including static and dynamic time reservations, are compared to ensure the availability of resources for high-priority patients upon their arrival. The success of the CG method is measured against actual data from Iridium Netwerk, Belgium's leading oncology centre. The results show that the dynamic time reservation strategy is the best way to deal with the unpredictability that arises from treating urgent patients. A sensitivity analysis demonstrates the robustness of this method against fluctuating arrival rates. Notably, the CG approach generates schedules meeting medical and technical constraints within reasonable computation times at Iridium Netwerk. In [9], the many optimization techniques that have been developed to address the RT patient scheduling problem are described and discussed at length. This study offers valuable insights for developing automated scheduling algorithms in practical healthcare settings. The paper introduces an Integer Programming (IP) model, a Column Generation IP model (CG-IP), and a Constraint Programming model, all designed to accommodate the scheduling of patients across various machine types while considering treatment priority, session duration, and machine availability. The models are enriched with expected future patient arrivals as placeholders. Multiple objective functions, including minimizing wait times and maximizing patient treatment time preferences, are employed to compare the models' performance. The results indicate that the CG-IP approach effectively manages diverse challenging scenarios within one hour, demonstrating an optimal performance gap of less than 1%. This underscores its potential as a valuable tool for automated RT scheduling. In [10], a data-driven approach to patient appointment scheduling is presented, addressing the challenges associated with patient punctuality. The study employs data mining techniques to predict patient behaviour, particularly regarding lateness and appointment misses, which are subsequently integrated into the Patient Appointment Scheduling (PAS) system. An innovative two-phase prioritization strategy is presented, taking into account not only the patient's health status improvement over time but also the possibility of treatment extension. A Mixed Integer Linear Programming (MILP) model is used to identify the ideal patient schedule for therapy, incorporating accommodations for no-show patients, based on the forecasts and prioritization. The developed technique is tested in both single and multi-server radiotherapy facilities, demonstrating a significant reduction in the total expense of the centre, up to 30%.

The research [11] provides a complete study of Appointment Scheduling in Medical Services. The study categorizes and examines over 150 scientific papers, highlighting various problem specifications and approaches in outpatient scheduling. The review categorizes methods for shortening patient waiting times and optimizing healthcare centre operations, emphasizing practical techniques such as mathematical optimization, simulation models, Markov chains, and artificial intelligence. The paper aims to offer insights into outpatient scheduling problems at different operational levels and provides an overview of hybrid modelling approaches. In [12], a novel two-stage appointment scheduling problem is formulated based on historical data from a tertiary cancer centre. Patients are categorized, and the objective is to minimize wait times for standard patients while optimally utilizing departmental resources. The model considers pre-treatment times and machine utilization forecasts. Real-world patient data is used to validate the model's performance, showcasing accurate predictions of utilization and potential reductions in wait times. The study presented in [13] focuses on implementing Operations Research (OR) techniques for generating RT patient schedules in a clinical setting. The procedure used in the OR is meant to reflect clinical scheduling goals while also satisfying technical and medical constraints. Patients waiting time, treatment time uniformity, and equipment usage all increase significantly when computerized schedules are compared to those prepared manually in the past. This automated approach also offers substantial time savings in administrative work. The research [14] proposes a two-phase method to the Radiotherapy Scheduling Issues. Employing Integer Linear Programming, the first step allocates sessions to particular linear acceleration devices and days. The second phase determines patient sequences and appointment times using MILP and Constraint Programming. Real data from a large cancer centre in Canada is used for testing. The results indicate that the approach effectively reduces waiting times, with Constraint Programming excelling in quickly finding good solutions and MILP excelling in closing optimality gaps.

## III. THEORETICAL CONCEPT

### A. Genetic Algorithm

A GA is a heuristic optimization strategy motivated by natural selection [15]. Complex optimization and search challenges can be tackled with its help [16]. In the context of patient scheduling in radiotherapy, a GA can be employed to find an optimal schedule for treating patients while considering various factors like patient waiting time, treatment time, and resource utilization. Here's how the GA process works in the context of patient scheduling for radiotherapy:

1. Chromosome Representation: In this study, the chromosome is a data structure that encodes the information needed for patient scheduling. It consists of several components: patient awaiting scheduling, the range of potential future scheduling scenarios, the number of patients within each scenario, the initiation day of treatment, and the available time slots designated for treatment sessions.

2. Chromosome Structure: The chromosome is divided into two equal parts. The first part encodes the patient's day of beginning treatment, while the second part encodes the time slot for beginning treatment. The arrangement of genes (sections of the chromosome) is crucial, as the position of a gene determines whether it encodes the beginning day or the time slot for treatment.

3. Population Initialization: The GA starts by generating an initial population of chromosomes [17]. Each chromosome represents a potential solution (patient schedule) to the problem. The chromosomes are created by encoding patients' information as described above.

4. Fitness Function: A fitness function evaluates how well a particular chromosome (patient schedule) performs with respect to the optimization goals. In the context of patient scheduling, the fitness function could consider factors like minimizing patient waiting times, optimizing resource utilization, and balancing treatment schedules.

5. Selection: The selection process simulates the principle of "survival of the fittest" [18]. Chromosomes with higher fitness scores have a better chance of being selected for reproduction. The roulette wheel selection is employed.

6. Crossover: Crossover entails merging genetic material from two parent chromosomes to generate one or more offspring chromosomes. In this context, it could mean exchanging information about beginning days and time slots between parents to create new potential schedules.

7. Mutation: Mutation introduces minor, random alterations in chromosomes to uphold genetic diversity within the population. In the context of patient scheduling, this could involve changing a patient's treatment day or time slot randomly.

8. Decoding: After undergoing crossover and mutation, the chromosomes are decoded to extract the actual patient scheduling information. This information includes session details, patient type, chosen time slot, and probability of scenarios.

9. Termination: The GA proceeds by repetitively navigating the steps of selection, crossover, and mutation for a defined number of generations or until a specific stopping condition is fulfilled. This stopping condition might involve reaching a maximum iteration count or identifying a satisfactory solution. Upon the culmination of the GA process, the finest chromosome (schedule) within the concluding population is extracted as the optimal resolution to the patient scheduling predicament.

*B. FireFly Optimization*

The Firefly Algorithm (FA) is a nature-inspired optimization technique that mimics the flashing behaviour of fireflies to find optimal solutions in complex optimization problems [19]. When applied to patient scheduling in radiotherapy, FA aims to determine the best arrangement of patient treatments to minimize waiting times, optimize resource usage, and adhere to treatment constraints. Firefly Algorithm Steps for Patient Scheduling:

1. Initialization: At the outset, a population of fireflies is created [20]. Each firefly embodies a potential scheduling solution, corresponding to a unique arrangement of patients' treatment schedules. These fireflies are positioned in a search space that represents different scheduling possibilities.

2. Fitness Evaluation: For each firefly, a fitness function is defined to quantify the quality of a scheduling solution. This function considers aspects such as minimizing patient waiting times, optimizing resource utilization, and satisfying treatment requirements. The higher the fitness score, the better the scheduling solution.

3. Light Intensity Assignment: To simulate the flashing behaviour of fireflies, light intensity values are assigned to each firefly based on their fitness scores [21]. Fireflies with higher fitness scores emit brighter light, indicating their proximity to an optimal solution.

4. Movement and Attraction Mechanism: Fireflies are attracted to brighter fireflies within the search space, analogous to fireflies being drawn to brighter lights in nature. The attractiveness between two fireflies is determined by a function that factors in their relative brightness and the distance between them. This attraction guides fireflies toward more promising solutions. The attractiveness between two fireflies i and j can be computed as:

$$Attractiveness(i,j) = \beta * exp(-\gamma * Distance(i,j)^2) \quad [1]$$

where $\beta$ and $\gamma$ are control parameters.

5. Position Update: Each firefly updates its position based on the attractiveness of other fireflies. This entails calculating a new position that balances the effect of the attraction and incorporates randomness to explore uncharted territories in the search space.

$$New\ Position = Current\ Position + Attractiveness * (Brighter\ Firefly's\ Position - Current\ Position) + Random\ Component \quad [2]$$

6. Light Intensity Update: After position updates, fireflies reassess their light intensities based on their new positions and fitness values. The light intensity is adjusted to align with the improved fitness scores, effectively signaling better solutions.

$$New\ Light\ Intensity = Initial\ Light\ Intensity + \alpha * Fitness\ Value \quad [3]$$

7. Termination Criteria: The algorithm runs for a predefined number of iterations or until a specified stopping criterion is met. This criterion could be the attainment of a satisfactory solution or convergence of the fitness values to a stable state. Once the algorithm concludes, the firefly with the highest light intensity—indicating the best fitness score—is selected as the optimized solution. This solution encapsulates a patient scheduling arrangement that minimizes waiting times, optimizes resource utilization, and adheres to treatment constraints.

*C. Wolf Optimization*

The WO Algorithm is a nature-inspired optimization technique that simulates the hunting behaviour of wolf packs to solve optimization problems [22-24]. Here's how it can be applied to patient scheduling. Steps of the WO Algorithm:

1. Initialization: At the start, a population of wolf packs is formed, where each pack embodies a potential scheduling solution [25]. The position of each wolf represents a unique arrangement of patient treatment schedules. The chromosome structure contains the patient's beginning treatment day and the time slot of treatment initiation.

2. Fitness Function: A fitness function is formulated to gauge the quality of each scheduling solution. This function takes into account various factors such as minimizing patient waiting times, optimizing resource utilization, adhering to treatment constraints, and considering the specified details of the problem statement. The higher the fitness value, the better the solution.

3.Pack Hierarchy: In the hierarchy of wolf packs, the alpha wolf serves as the leader, beta wolves are followers, and omega wolves are the least dominant [26]. This hierarchy imitates the leadership structure in wolf packs. The alpha wolf explores new regions, beta wolves follow, and omega wolves contribute to diversified exploration.

4.Leader Wolf Movement: The alpha wolf simulates exploration, seeking new potential solutions. It adjusts its position based on a random coefficient $A$ and the distance $D$ between the alpha wolf and the prey (target solution). The random coefficient A controls the step size of the exploration.

$$Alpha\_new = Alpha - A * D \quad [4]$$

5.Prey Wolf Movement: Beta wolves play a crucial role in exploiting known solutions. They adjust their positions by averaging the positions of the alpha and beta wolves. This collaborative movement towards the alpha wolf fosters convergence toward promising regions.

$$Beta_{new} = \frac{(Alpha + Beta)}{2} \quad [5]$$

6.Omega Wolf Movement: Omega wolves, representing the least dominant pack members, contribute to diversified exploration. They adjust their positions using the average of the alpha and omega positions. This movement introduces diversity into the search space.

$$Omega_{new} = \frac{(Alpha + Omega)}{2} \quad [6]$$

7.Convergence and Exploration: The wolf hierarchy and their respective movements balance convergence and exploration. Alpha wolves lead exploration, beta wolves' focus on exploiting known solutions, and omega wolves diversify the search space. This collective behaviour facilitates the algorithm's ability to find optimal solutions.

8.Termination Criteria: The algorithm stops when a predefined termination condition is met. This condition can be a maximum number of iterations or when fitness values converge within a specific threshold. Upon termination, the positions of the alpha, beta, and omega wolves are analysed. The wolf with the best fitness value is selected as the optimal solution. Decoding the position reveals the patient's session, type, chosen slot, and scenario probability.

9.

## IV. OBJECTIVE FUNCTIONS

As patients increasingly seek shorter waiting times for enhanced treatment and recovery, the typical procedure in many countries involves treatment planners receiving the doctor's treatment plan and then scheduling patients based on their availability, confirmed through phone calls. Notably, certain patients have specific time preferences for their treatment slots, while others possess flexibility in timing. Thus, this study segregates patients into two categories: general patients, open to various time slots, and special patients, constrained to specific time slots. As an example, certain patients may exclusively be available for RT during the afternoon hours due to prior commitments in the forenoon. When scheduling, prioritizing the needs of regular patients over those of special patients can cause the latter to wait longer than necessary and could harm their health.

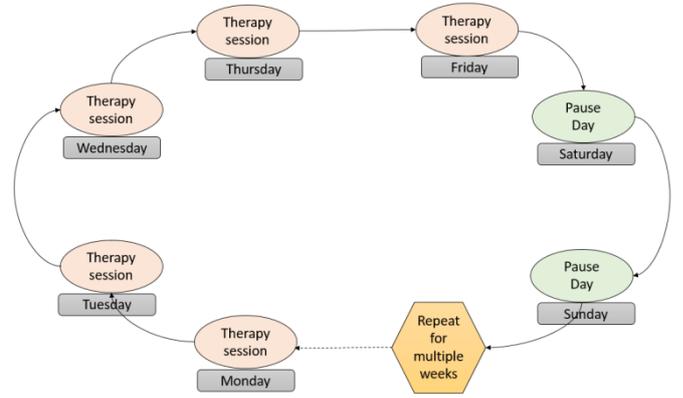

Fig. 1. Radiotherapy Schedule

To avert this situation and diminish patient wait times, various scenarios are contemplated and integrated into stochastic optimization. The research's underlying assumptions are as follows:

- Radiotherapy services are provided from Monday to Friday within a week. Figure 1 illustrates a standard fractionation scheme for a patient undergoing RT.
- Subsequent radiotherapy sessions must be conducted on consecutive working days after the initial treatment.
- One patient is treated at a time by each machine.
- The order and time taken for operations are irrelevant.
- Device setup time and inter-machine transfers are negligible.
- Machines remain available throughout the shifts.
- Once treatment commences, the assigned machine cannot be changed.
- All devices are identical.
- Adequate labor resources are consistently available.
- Treatment planning includes machine selection.
- All patients are considered standard cases.

The main aim of this study is to reduce the collective waiting time experienced by patients. The objective function is divided into offline, online, and OS (Online Stochastic).

### A. Offline objective function

In addressing the offline scheduling issue, we have established a finite scheduling duration, assuming a predetermined number of days. We operate under the premise that all patient requests are already available beforehand. Consequently, appointment determinations can be made by taking into account the complete set of patient information.

Let $J$ denote the index set of patients. We define $J_s$ and $J_g$ as the sets of special and general patients, respectively. $J^w$, $J_s^w$, and $J_g^w$ represent the patient set, special patient set, and general patient set for scenario $\omega$, where $\omega$ belongs to $\Omega$, the scenario space. Here, $i$ from $I$ signifies the selected machines for treating patient $j$. Similarly, $t$ from $T$ corresponds to time slots, $l$ indicates the therapy day for a patient, and $L$ designates the scheduling day. $S_{js} \epsilon T$ signifies the set of chosen treatment time slots for patients in $j\epsilon J_s$, while $S_{js}^w \in T$ indicates the set of selected treatment time slots for patients in $j\epsilon J_s$ within scenario $\omega$. $p_j$ refers to the total treatment sessions for patient $j$, and $p_j^\omega$ is the total treatment sessions for patient $j$ within scenario $\omega$. The variable $F_{ilt}$ is assigned a value of 1 when machine $i$ is available for scheduling during time slot t on day l, and it takes a value of 0 otherwise. The decision variables are represented as $X_{ijlt}$ and $Y_{ij'lt}^\omega$. If patient $j$ is slotted to utilize machine $i$ at time slot $t$ on treatment day l, $X_{ijlt}$ is set to 1; otherwise, it remains 0. $Y_{ij'lt}^\omega$ is designated as 1 if patient $j'$ is scheduled to use machine $i$ at time slot $t$ on treatment day l within scenario $\omega$; otherwise, it's set to 0.

Offline objective function
$$Min \sum_{j \in J} \sum_{l \in L} \sum_{i \in I} \sum_{t \in T} l * X_{ijlt} \quad [7]$$
Constraints
$$\sum_{j \in J} X_{ijlt} \leq 1, i \in I, l \in L, t \in T \quad [7a]$$
$$X_{ijlt} \leq F_{ilt}, i \in I, j \in J, l \in L, t \in T \quad [7b]$$
$$\sum_{l \in L} \sum_{t \in S_{js}} X_{ijlt} = p_j, i \in I, j \in J_s \quad [7c]$$
$$\sum_{l \in L} \sum_{t \in T} X_{ijlt} = p_j, i \in I, j \in J_g \quad [7d]$$
$$\sum_{t \in T} \sum_{l=0}^{L+1} |X_{ijlt} - X_{ij(l-1)t}| = 2, i \in I, j \in J \quad [7e]$$
$$X_{ijlt} \in \{0,1\}, i \in I, j \in J, l \in L, t \in T \quad [7f]$$

The objective function (7) seeks to minimize the total treatment days, influenced by the initial treatment day and the patient's duration, which is pre-determined upon reservation. Hence, this objective function reflects the degree of minimal total patient waiting time. According to Equation (7a), it is possible to schedule just one patient for machine $i$ at the period $t$ on day $l$. By solving (7b), we know that on that particular day, the capacity of the device $i$ won't be surpassed. Equations (7c) and (7d) ascertain the scheduling of patients, particularly those in sets $j \epsilon J_s$ and $j \epsilon J_g$, within time slots $t \in S_{js}$ and $t \in T$ respectively, on day l, with respect to machine $i$. Equation (7e) mandates a continuous treatment experience for patients allocated to machine $i$ during time slot $t$ on day l, encompassing their entire treatment session. In cases where $l$ equals 0 or $L + 1$, both decision variables $X_{ij0t}$ and $X_{ij(L+1)t}$ are defined as 0. Finally, Equation (7f) establishes binary conditions for the decision variables.

*B. Online objective function*

In the context of online RT scheduling, patient requests aren't predetermined; instead, they arrive sequentially, one by one. Appointments are scheduled as requests come in. A number of patients have been scheduled in advance of time t, therefore let's say $j_t$ represents a patient's request at time $t$. The goal here is to schedule patient $j_t$ while keeping the appointments of previously scheduled patients unchanged. The objective function, as well as the decision variables and constraints, remains consistent with those of the offline scheduling problem (constraints 7a to 7f).

$$Min \sum_{l \in L} \sum_{i \in I} \sum_{t \in T} l * X_{ijlt} \quad [8]$$

*C. OS Objective Function*

The OS problem constitutes a part of the broader online objective functions. Here our focus shifts to scheduling patient requests as they arrive while keeping other patients' schedules unchanged. Furthermore, we take into consideration future patient requests while making these appointments. Similar to the offline objective function (7a)-(7f), the requirements for the OS problem are also consistent. However, the objective function (9) takes on an alternate form. This function incorporates an additional factor that considers the anticipated result of a subfunction, alongside the aim of minimizing the current patient's waiting time. Each of the (9a)–(9g) subfunctions represents a unique "scenario $\omega$" for the issue. These subfunctions, like the offline function, use the parameterized value of $X_{ijlt}$ t to determine the value of the decision variable $Y_{ij'lt}^\omega$. $Y_{ijlt}^\omega$ matches $X_{ijlt}$, and both $Y_{ij'0t}^\omega$ and $Y_{ij'(L+1)t}^\omega$ are defined as 0.

$$Min \sum_{l \in L} \sum_{i \in I} \sum_{t \in T} l * X_{ijlt} + E(Q(x, \omega)) \quad [9]$$

Mathematical Model for scenario $\omega$
$$Q(x, \omega) Min \sum_{j \in J} \sum_{l \in L} \sum_{i \in I} \sum_{t \in T} l * Y_{ij'lt}^\omega \quad [9a]$$

Constraints
$$X_{ijlt} + \sum_{j' \in j^\infty} Y_{ij'lt}^\omega \leq 1, i \in I, l \in L, j \in J, t \in T, \omega \in \Omega \quad [9b]$$
$$X_{ijlt} + Y_{ij'lt}^\omega \leq F_{ilt}, i \in I, l \in L, j' \in J^\omega, t \in T, \omega \in \Omega \quad [9c]$$
$$\sum_{l \in L} \sum_{t \in S_{j's}^\omega} Y_{ij'lt}^\omega = P_{j'}^\omega, i \in I, j' \in J_s^\omega, \omega \in \Omega \quad [9d]$$
$$\sum_{l \in L} \sum_{t \in T} Y_{ij'lt}^\omega = P_{j'}^\omega, i \in I, j' \in J_s^\omega, \omega \in \Omega \quad [9e]$$
$$\sum_{t \in T} \sum_{l=0}^{L+1} |Y_{ij'lt}^\omega - Y_{ij'(l-1)t}^\omega| = 2, i \in I, j' \in J_s^\omega, \omega \in \Omega \quad [9f]$$
$$X_{ijlt}, Y_{ij'lt}^\omega \in \{0,1\}, i \in I, l \in L, j' \in J, t \in T, \omega \in \Omega \quad [9g]$$

V. RESULT AND DISCUSSION

In this section, we delve into the detailed outcomes of our study. The performance evaluation is done using objective values, runtime, and convergence time. To comprehensively evaluate the performance of our optimization algorithms in diverse scenarios, we meticulously designed a series of experiments comprising a total of 10 cases. These cases were then artfully grouped into three distinct scenarios: involving 4, 8, and 10 cases respectively. The algorithms chosen for our investigation were the GA, FFO, and WO.

*A. Evaluation of objective values*

To gauge the effectiveness of these algorithms, we examined several critical metrics. These metrics included the average, best, and worst objective values for each algorithm. In Figure 2.a, we depict the assessment of objective values for four cases. Additionally, Figures 2.b and 2.c provide a visualization of objective value evaluations encompassing eight and ten cases, respectively. Our observations, presented graphically in Figure 2, undeniably highlight the consistent supremacy of WO across all scenarios. Notably, WO consistently outperformed both GA and FFO in terms of good, average, and poor objective values. This suggests that WO possesses remarkable search capabilities, adept at approaching solutions that are near-optimal. Furthermore, the remarkable narrowness of the standard deviation and confidence interval associated with WO attests to its heightened reliability.

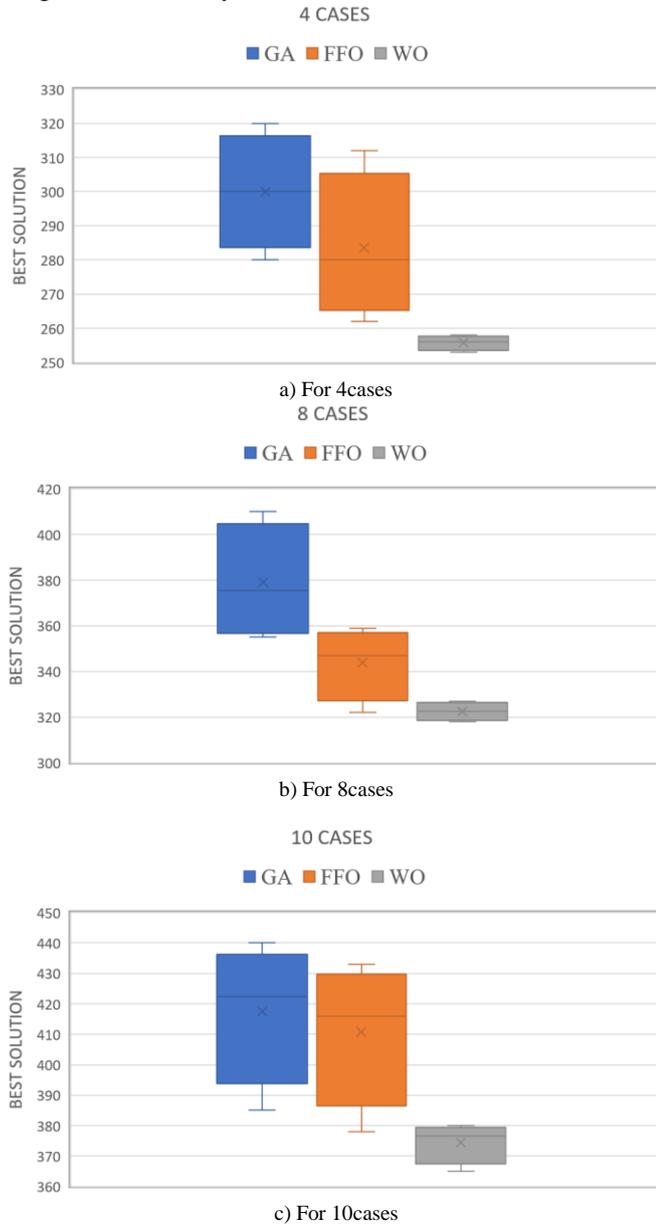

Fig. 2. Comparison of objective values

### B. Evaluation of runtime

Shifting our focus to runtime analysis, we diligently recorded the execution times for GA, FFO, and WO across various case sizes and the obtained result is given in table 1. For instance, when considering 4 cases, GA exhibited an execution time of 5.87 seconds, while FFO took 6.78 seconds, and WO impressively completed in 5.65 seconds. Similarly, for 8 cases, GA required 8.75 seconds, FFO took 9.52 seconds, and WO's runtime was 11.07 seconds. As we extended our analysis to 10 cases, GA's runtime increased to 9.55 seconds, FFO required 11.07 seconds, and WO efficiently concluded in 9.12 seconds. Figure 3 graphically portrays the average runtime across different scenarios. Here, WO emerges as the most efficient algorithm for tests involving 4, 8, and 10 cases. Importantly, it is worth noting that while there are discernible variations in runtime among the algorithms, these differences generally do not reach levels of substantial significance.

Table 1. Evaluation of runtime

| Algorithm | 4 cases | 8 cases | 10 cases |
|---|---|---|---|
| GA | 5.87 | 8.75 | 9.55 |
| FFO | 6.78 | 9.52 | 11.07 |
| WO | 5.65 | 8.43 | 9.12 |

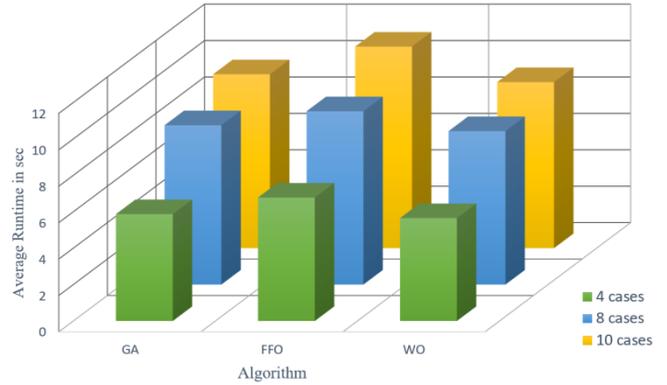

Fig. 3. Comparison of running time

### C. Evaluation of convergence behaviour

Intriguingly, the convergence behaviour of each algorithm is displayed in Figure 4. Figures 4.a, 4.b, and 4.c showcase the convergence behaviour for different scenarios including 4, 8, and 10 cases, respectively. This visualization unequivocally confirms WO's outstanding performance in terms of convergence. On the other hand, GA exhibits relatively modest convergence capabilities. FFO, while generally superior to WO in terms of convergence, still showcases convergence that is not substantially different from WO.

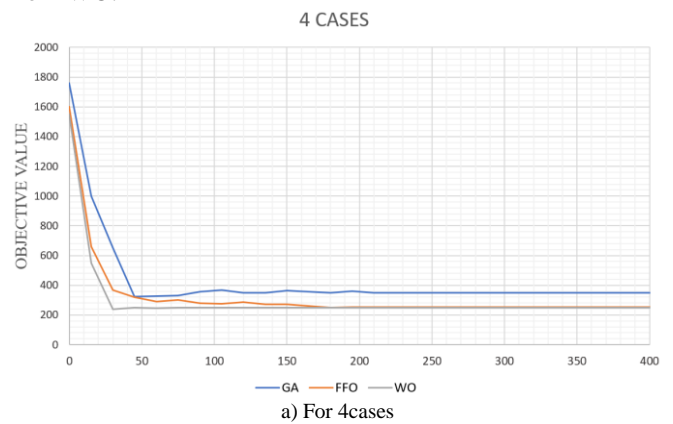

a) For 4cases

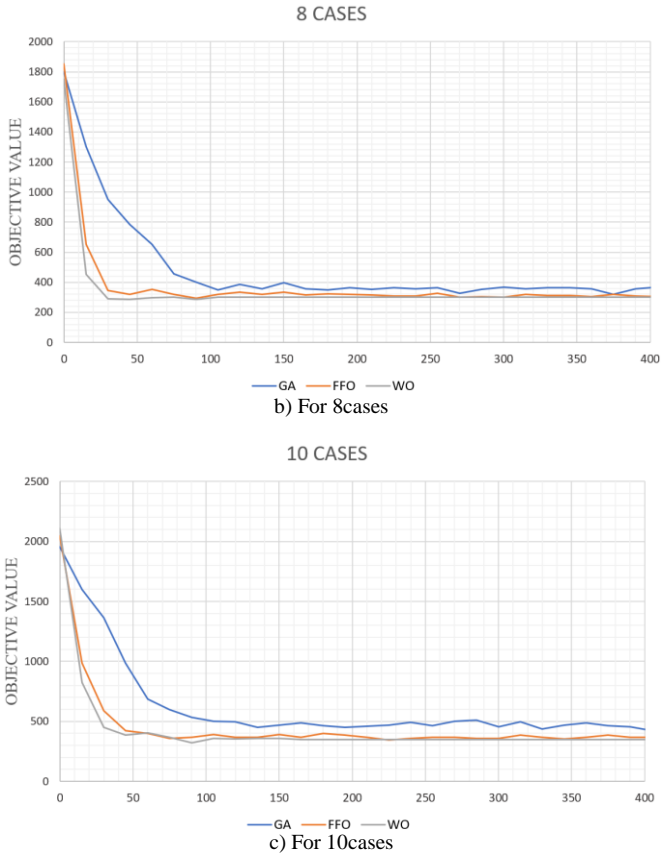

b) For 8cases

c) For 10cases

Fig. 4. Convergence Condition of various scenario

### D. Comparison of scheduling outcome

By examining the outcomes presented in Table 2, a comparative analysis was conducted between the outcomes of the current scheduling method and the optimized scheduling approach. The findings in the table unequivocally indicate that the scheduling achieved through the WO method surpasses both the present scheduling approach and alternative optimization methods in terms of performance.

Table 2. Comparison of present and OS scheduling

| Algorithm | Waiting Days | Waiting Patients |
|---|---|---|
| Manual | 47 | 8 |
| GA | 32 | 5 |
| FFO | 25 | 5 |
| WO | 21 | 3 |

### VI. CONCLUSION

In light of the escalating prevalence of cancer, there has been a corresponding surge in demand for RT. As the need for RT continues to rise, it becomes increasingly imperative to optimize the allocation of resources within this critical domain. To this end, the application of algorithms to automatically generate patient schedules emerges as a pivotal strategy, replacing the current manual practices prevalent in most medical clinics. This paper stands as a pivotal contribution, serving as a decision support tool for the practical implementation of scheduling algorithms. The study's distinctive focal point is the introduction and thorough evaluation of the Optimization algorithm. From convergence time to runtime and objective values, WO consistently outperforms its counterparts, reaffirming its effectiveness in solving the intricate scheduling challenges in RT. Furthermore, the computational findings underscore the potential impact of optimized scheduling on patient experiences, demonstrated by a reduction in waiting times. Moving forward, expanding the scope of bio-inspired algorithms to include newer and more innovative approaches can lead to even more optimized and efficient patient scheduling solutions. Moreover, exploring the potential integration of machine learning and artificial intelligence techniques could contribute to enhancing the accuracy and adaptability of the scheduling process. In summary, this study lays a strong foundation for the implementation of biomimicry-driven optimization in RT patient scheduling.

REFERENCE

1. Weiderpass, Elisabete, and BERNARD W. Stewart. "World cancer report." *The Int. Agency for Res. on Cancer (IARC)* (2020).
2. Braune, Roland, Walter J. Gutjahr, and Petra Vogl. "Stochastic radiotherapy appointment scheduling." *Central European Journal of Operations Research* (2021): 1-39.
3. Gupta D, Denton B (2008) Appointment scheduling in health care: challenges and opportunities. IIE Trans 40:800–819.
4. Kazemian P, Sir MY, Van Oyen MP, Lovely JK, Larson DW, Pasupathy KS. Coordinating clinic and surgery appointments to meet access service levels for elective surgery. J Biomed Inform. 2017 Feb;66:105-115.
5. Kuiper, Alex, Michel Mandjes, Jeroen de Mast, and Ruben Brokkelkamp. "A flexible and optimal approach for appointment scheduling in healthcare." *Decision Sciences* 54, no. 1 (2023): 85-100.
6. Mandelbaum, Avishai, Petar Momčilović, Nikolaos Trichakis, Sarah Kadish, Ryan Leib, and Craig A. Bunnell. "Data-driven appointment-scheduling under uncertainty: The case of an infusion unit in a cancer center." *Management Science* 66, no. 1 (2020): 243-270.
7. Mackillop WJ. Killing time: the consequences of delays in radiotherapy. Radiother Oncol. 2007 Jul;84(1):1-4. Epub 2007 Jun 14. PMID: 17574695.
8. Frimodig, Sara, Per Enqvist, and Jan Kronqvist. "A Column Generation Approach for Radiation Therapy Patient Scheduling with Planned Machine Unavailability and Uncertain Future Arrivals." arXiv preprint arXiv:2303.10985 (2023).
9. Frimodig, Sara, Per Enqvist, Mats Carlsson, and Carole Mercier. "Comparing optimization methods for radiation therapy patient scheduling using different objectives." arXiv preprint saraarXiv:2211.01150 (2022).
10. Moradi, Shahryar, Mehdi Najafi, Sara Mesgari, and Hossein Zolfagharinia. "The utilization of patients' information to improve the performance of radiotherapy centers: A data-driven approach." Computers & Industrial Engineering 172 (2022): 108547.
11. Ala, Ali, and Feng Chen. "Appointment scheduling problem in complexity systems of the healthcare services: A comprehensive review." Journal of Healthcare Engineering 2022 (2022).
12. Jia, Fan, Michael Carter, and Srinivas Raman. "Data-Driven Two-stage Appointment Radiotherapy Scheduling Model for Resource Optimization at a Tertiary Cancer Center." (2023).
13. Frimodig, Sara, Carole Mercier, and Geert De Kerf. "Automated Radiation Therapy Patient Scheduling: A Case Study at a Belgian Hospital." arXiv preprint arXiv:2303.12494 (2023).
14. Pham, Tu-San, Louis-Martin Rousseau, and Patrick De Causmaecker. "A two-phase approach for the Radiotherapy Scheduling Problem." *Health Care Management Science* (2022): 1-17.
15. SS, Vinod Chandra, and Anand HS. "Nature inspired meta heuristic algorithms for optimization problems." *Computing* 104, no. 2 (2022): 251-269.
16. Alam, Tanweer, Shamimul Qamar, Amit Dixit, and Mohamed Benaida. "Genetic algorithm: Reviews, implementations, and applications." *arXiv preprint arXiv:2007.12673* (2020).
17. Immanuel, Savio D., and Udit Kr Chakraborty. "Genetic algorithm: An approach on optimization." In *2019 international conference on communication and electronics systems (ICCES)*, pp. 701-708. IEEE, 2019.
18. Brouwer, Nielis, Danny Dijkzeul, Levi Koppenhol, Iris Pijning, and Daan Van den Berg. "Survivor selection in a crossoverless evolutionary algorithm." In *Proceedings of the Genetic and Evolutionary Computation Conference Companion*, pp. 1631-1639. 2022.

19. Ghasemi, Mojtaba, Soleiman kadkhoda Mohammadi, Mohsen Zare, Seyedali Mirjalili, Milad Gil, and Rasul Hemmati. "A new firefly algorithm with improved global exploration and convergence with application to engineering optimization." *Decision Analytics Journal* 5 (2022): 100125.
20. Azizan, Muhammad Naqiuddin Irfan Heiqal, and See Pheng Hang. "Investigation of Firefly Algorithm using Multiple Test Functions for Optimization Problem."
21. Alhadawi, Hussam S., Dragan Lambić, Mohamad Fadli Zolkipli, and Musheer Ahmad. "Globalized firefly algorithm and chaos for designing substitution box." *Journal of Information Security and Applications* 55 (2020): 102671.
22. Sharma, Satyam, Ridhi Kapoor, and Sanjeev Dhiman. "A novel hybrid metaheuristic based on augmented grey wolf optimizer and cuckoo search for global optimization." In *2021 2nd International Conference on Secure Cyber Computing and Communications (ICSCCC)*, pp. 376-381. IEEE, 2021.
23. Hou, Yuxiang, Huanbing Gao, Zijian Wang, and Chuansheng Du. "Improved grey wolf optimization algorithm and application." *Sensors* 22, no. 10 (2022): 3810.
24. Nadimi-Shahraki, Mohammad H., Shokooh Taghian, and Seyedali Mirjalili. "An improved grey wolf optimizer for solving engineering problems." *Expert Systems with Applications* 166 (2021): 113917.
25. Abed-Alguni, Bilal H., and Noor Aldeen Alawad. "Distributed Grey Wolf Optimizer for scheduling of workflow applications in cloud environments." *Applied Soft Computing* 102 (2021): 107113.
26. Sharma, Isha, Vijay Kumar, and Sanjeewani Sharma. "A comprehensive survey on grey wolf optimization." *Recent Advances in Computer Science and Communications (Formerly: Recent Patents on Computer Science)* 15, no. 3 (2022): 323-333.